\documentclass[10pt,twocolumn,letterpaper]{article}

\usepackage[pagenumbers]{cvpr} 

\usepackage{graphicx}
\usepackage{amsmath}
\usepackage{amssymb}
\usepackage{booktabs}
\usepackage{multirow}
\usepackage{xcolor}

\usepackage{color}
\definecolor{darkgreen}{RGB}{30,150,30}
\definecolor{darkblue}{RGB}{0,0,127}
\definecolor{darkred}{RGB}{180,20,20}

\usepackage[pagebackref,breaklinks,colorlinks]{hyperref}

\hypersetup{
  linkcolor    = darkred, 
  citecolor   = darkgreen 
}

\usepackage{caption}
\usepackage[capitalize]{cleveref}
\crefname{section}{Sec.}{Secs.}
\Crefname{section}{Section}{Sections}
\Crefname{table}{Table}{Tables}
\crefname{table}{Tab.}{Tabs.}

\newcommand{\vect}[1]{\mathbf{#1}}

\def\cvprPaperID{4274} 
\def\confName{CVPR}
\def\confYear{2022}

\begin{document}

\title{Sparse Fusion for Multimodal Transformers}

\author{Yi Ding\thanks{Equal contribution} \! \footnote , \quad Alex Rich\footnotemark[1]  \footnotemark[2]  \quad Mason Wang\footnote \!  \quad Noah Stier\footnotemark[2] \\  \quad Matthew Turk\footnote \! \quad Pradeep Sen\footnotemark[2] \quad Tobias H\"ollerer\footnotemark[2]\\
\vspace{-.3cm}\\

\footnotemark[2] ~University of California, Santa Barbara\\
\footnotemark[3] ~Saratoga High School \\
\footnotemark[4] ~Toyota Technological Institute at Chicago\\
{\tt\small yding@ucsb.edu, anrich@ucsb.edu, mason.wang0025@gmail.com, noahstier@ucsb.edu,} \\ 
{\tt\small  mturk@ttic.edu, psen@ece.ucsb.edu, holl@ucsb.edu}
}
\maketitle

\begin{abstract}
Multimodal classification is a core task in human-centric machine learning.
We observe that information is highly complementary across modalities, thus unimodal information can be drastically sparsified prior to multimodal fusion without loss of accuracy.
To this end, we present Sparse Fusion Transformers (SFT), a novel multimodal fusion method for transformers that performs comparably to existing state-of-the-art methods while having greatly reduced memory footprint and computation cost. 
Key to our idea is a sparse-pooling block that reduces unimodal token sets prior to cross-modality modeling.
Evaluations are conducted on multiple multimodal benchmark datasets for a wide range of classification tasks. State-of-the-art performance is obtained on multiple benchmarks under similar experiment conditions, while reporting up to six-fold reduction in computational cost and memory requirements. Extensive ablation studies showcase our benefits of combining sparsification and multimodal learning over naive approaches. This paves the way for enabling multimodal learning on low-resource devices.
\end{abstract}

\section{Introduction}

We experience and interact with the world through our five senses: sight, sound, taste, touch, and smell. 
The human brain is incredibly good at processing all of this information, paying attention only to the few things that matter. 
Imbuing a computer with the ability to process multimodal data effectively is highly desirable because it would enable a vast array of multi-sensory applications. 
However, processing multiple data streams increases computational cost, and it is therefore a high priority to develop efficient algorithms in this domain.
Additionally, many of these applications, such as the detection of instances of domestic abuse, or detection of prolonged emotional and psychological struggles, are particularly well-suited for mobile or low-resource devices. 
In these resource-constrained settings, the computation cost and memory footprint become critical factors that must be considered for practical use.

Current multimodal algorithms involve some level of modality-independent feature processing followed by a fusion process which then jointly models the dependencies and cross-dependencies between the modalities. In particular, deep-learning transformer models have been used in this way to achieve state-of-the-art performance on numerous tasks~\cite{rahman-etal-2020-integrating, nagrani2021attention}. However training and processing such data remains prohibitively expensive in many cases, in terms of time, computational resources, and energy consumption. For example, a single layer of a vision transformer \cite{dosovitskiy2020vit} requires approximately 1.35 billion floating-point operations (GFlops) for a $224 \times 224$ image for a single forward pass. If we represent a sequence of 30 frames in a similar manner for video data, this explodes to 88.24 GFlops. Although recent advancements have been made to sparsify transformers, these efforts have primarily approached the problem from a unimodal perspective~\cite{deitplmr, DBLP:journals/corr/abs-2104-03602, bao2021beit, pan2021scalable, wang2021not}.

Motivated by these concerns, we propose a sparse fusion method for multimodal transformers called Sparse Fusion Transformers (SFTs) that drastically reduces training time and memory consumption while maintaining the quality of existing fusion methods. 
Our approach is based on the hypothesis that the large amount of complementary information across different modalities allows us to sparsify unimodal information prior to multimodal fusion without the loss of accuracy.
In particular, approaching a problem from a multimodal perspective enables us to sparsify the unimodal information far more aggressively. 
With our sparse-fusion method, we achieve faster performance with less memory use while attending to features that are most important.

Our proposed fusion process is agnostic to input modality and makes a full multimodal classification network robust to sparsification of input representations. It is composed of three parts: a block-sparse within-modality attention to learn strong local representations, a pooling method for extracting them, and dense self-attention for cross-modal feature fusion. Furthermore, we propose to use a customized mixup to apply spatio-temporal regularization to the learned representations in a modality agnostic manner. Fusing features in this way demonstrates comparable or better performance than existing methods while requiring significantly less computation and memory. In summary, our contributions are:

\begin{itemize}
    \item We propose a novel fusion method that maintains or exceeds the performance of previous fusion methods while demonstrating up to a six-fold reduction in computation and memory requirements.
    \item We demonstrate that multimodal algorithms can tolerate far more token reduction than unimodal algorithms due to complementary cross-modal information. We show that by accounting for multimodal information during sparsification, more information can be removed without loss of performance. 
    \item We perform extensive ablation studies on fusion components using real-world datasets to determine the efficacy of each model component. We further experiment with multiple pooling methods to demonstrate model robustness under different pooling requirements.
\end{itemize}
\section{Related Work}

The problem of modality fusion has been explored in numerous problem spaces for a long time \cite{baltruvsaitis2018multimodal}. The primary challenge is to find an effective way to combine representations of data from disparate modalities into a single representation for more accurate modeling. While the first methods for multimodal fusion were proposed to address signal inadequacies in individual modalities \cite{yuhas1989}, we are now at a time when the resolution in each modality is much higher, making some computation costly and intractable. Therefore, we wish to purposely trade off some of the signal bandwidth to improve performance.  

Many methods have been proposed to tackle the task of fusion. A way to categorize all these techniques is by the time of fusion occurrence. Early fusion typically refers to combining base level representations or even input values, while late fusion primarily refers to its application near the output. Early deep-learning methods typically make use of linear layers and cross products to combine modalities \cite{wang2019words, zadeh2018memory, feichtenhofer2016convolutional}. More rudimentary forms of fusion simply involve adding the logits of individual modality predictions together. As transformer-based architectures have become very popular recently, some recent techniques have also explored their use in multimodal settings. Originally proposed in \cite{vaswani2017attention} for neural machine translation (NMT) tasks, they have demonstrated superior performance on multiple benchmark problems such as image classification \cite{dosovitskiy2020vit}, action recognition \cite{nagrani2021attention} and 3D reconstruction \cite{bozic2021transformerfusion, stier2021vortx}. The basic functionality is to apply layers of self-attention, on sequential representations. To classify a discrete output, transformers typically rely on the use of a special token (\texttt{CLS}) that is prepended to the sequence for classification.

The most natural form of transformer fusion is simply to concatenate the sequence of tokens and rely on self-attention to learn their inter-dependencies. Works such as \cite{tsai2019multimodal, jaegle2021perceiver} that do this learn better cross-modal representations and have shown benefits relative to naive fusion methods. Very recently, multimodal bottleneck transformers~\cite{nagrani2021attention} have demonstrated a way for early fusion to occur without the use of costly cross-modal operations. However, the process of fusing multimodal information with some form of concatenation and dense attention remains costly due to the $O(N^2)$ complexity of transformers for input sequences of length $N$. It is this cost we seek to address with our sparsification approach.

Recent efforts have focused on reducing computational complexity for transformers and large-scale deep learning~\cite{zhang2020accelerating, ren2021zero, rasley2020deepspeed, deitplmr}. An effective method for this is to exploit the representation of features within a small sliding window of tokens \cite{wang2019multi} on a long sequence. However, these methods require significant engineering efforts and are hard to train \cite{zaheer2020big}. Other works approach the problem via sparsification of the attention mechanism, such as random or local attention~\cite{kitaev2020reformer, rae2019compressive, ye2019bp}. Sparsification methods have also been applied successfully for some computer vision tasks \cite{pan2021scalable}.

Training optimizations for transformers have also been explored. Regularization techniques such as dropout\cite{srivastava2014dropout}, weight decay~\cite{loshchilov2018fixing}, and mixup~\cite{zhang2017mixup} have all been applied. While weight decay and dropout can be applied in a modality-agnostic manner directly onto the weights, the use of mixup has primarily been used to tackle problems in the vision domain, as its application is easily interpretable and offers large benefits to the algorithms \cite{nishi2021augmentation, berthelot2019mixmatch}. Although some recent efforts have been made to enable the application of mixup on domains in a modality agnostic manner \cite{verma2018manifold}, its application in a fundamentally multimodal domain remains underexplored. Its use in the mixing of fused features across modalities spatially and across time demonstrates large benefits for our application.


\section{Method}

\begin{figure}[t]
\centering
\includegraphics[width=.8\linewidth]{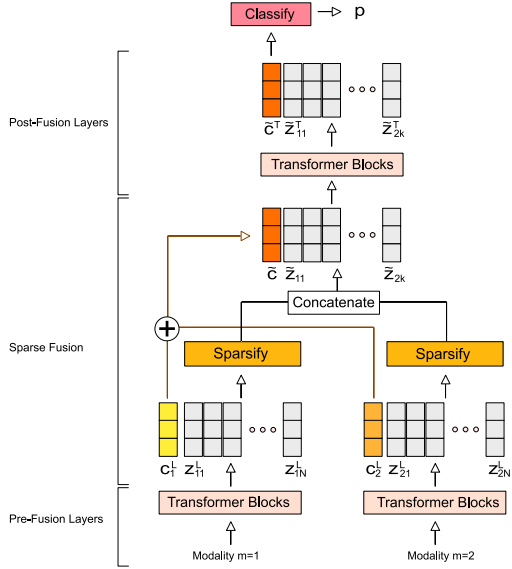}
\caption{Visualization of our fusion method with two modalities. Following existing work, a special \texttt{CLS} token is appended to each unimodal token set prior to unimodal transformers. After unimodal transformers, the \texttt{CLS} token ($\vect{c}_1^L$ and $\vect{c}_2^L$) from each modality is summed. A pooled block-sparse attention is applied to local regions of each modality. The \texttt{CLS} token and pooled representations are then combined, and dense self-attention is applied to model global and cross-modal dependencies.}
\label{fig:fusion}
\end{figure}

In this section, we describe our proposed Sparse Fusion Transformers (SFT).
See Fig.~\ref{fig:fusion} for a visualization of our algorithm.
As input, our method takes token sets from $M$ different modalities, $\vect{Z}_1, \dots, \vect{Z}_M$, with each modality consisting of $N$ tokens of dimension~$D$, $\vect{Z}_i~=~[\vect{z}_{i1}, \dots, \vect{z}_{iN}]~\in~\mathbb{R}^{N \times D}$.
Note the number of tokens $N$ can vary from modality to modality but for simplicity of notation, we keep it fixed in our description.
Additionally, if the token dimension $D$ varies from modality to modality, we apply a per-token projection to keep the token dimension constant across all modalities.
Following existing work, we prepend a special \texttt{CLS} token $\vect{c}$ with learnable parameters to each token set for each modality for the purpose of classification: $\hat{\vect{Z}}_i = [\vect{c}_i | \vect{Z}_i] = [\vect{c}_i, \vect{z}_{i1}, \dots, \vect{z}_{iN}] \in \mathbb{R}^{(N+1) \times D}$.
The goal of our method is classification, i.e., we want to learn a function $f_{\theta}:~\mathbb{R}^{M \times (N+1) \times D}~\rightarrow~\mathbb{R}^C$:
\begin{equation}
    f_{\theta}(\hat{\vect{Z}}_1, \dots, \hat{\vect{Z}}_M) = \vect{p},
\end{equation}
such that $\vect{p}$ is the probability distribution over $C$ classes.

Our method consists of three main parts.
First, we model relationships between tokens within modalities using a standard transformer that is applied unimodally (Sec.~\ref{sec:unimodal-trans}).
Second, we aggregate information within local regions of each sequence using block-sparse attention and then apply local subsequence pooling to sparsify the token set for each modality (Sec.~\ref{sec:fusion}).
Third, we concatenate the sparsified features from each modality and run dense self-attention to predict a final class (Sec.~\ref{sec:cross-modal}).
During training, we apply a novel multimodal variation of manifold mixup \cite{verma2018manifold} for regularization of intermediate latent representations (Sec.~\ref{sec:mixup}).

\subsection{Unimodal Modeling} \label{sec:unimodal-trans}
In this stage, we apply a separate transformer to the token set from each modality.
Following Vaswani \etal~\cite{vaswani2017attention}, we use a standard $L$-layer transformer encoder to model relationships between tokens in each modality. Each layer of the encoder consists of layer normalization (LN), Multi-head Self-Attention (MSA), and a Multi-Layer Perceptron (MLP).
Given token set $\hat{\vect{Z}}^l$ after $l$ transformer layers, the output of layer $l+1$ is:
\begin{equation}
    \vect{Y}^l = \mathrm{MSA}(\mathrm{LN}(\hat{\vect{Z}}^l)) + \hat{\vect{Z}}^l
\end{equation}
\begin{equation}
    \hat{\vect{Z}}^{l+1} = \mathrm{MLP}(\mathrm{LN}(\vect{Y}^l)) + \vect{Y}^l
\end{equation}
We apply a separate $L$-layer transformer per modality to get token sets $\hat{\vect{Z}}^L_1, \dots, \hat{\vect{Z}}^L_M$.

\subsection{Sparse Multimodal Fusion} \label{sec:fusion}
In this stage, we apply local pooling blocks to each token set $\vect{Z}^L_i$ to extract $k$ descriptive tokens per modality $\tilde{\vect{Z}}_i~=~[\tilde{\vect{z}}_{i1}, \dots, \tilde{\vect{z}}_{ik}]~\in~\mathbb{R}^{k \times D}$, as represented by the ``Sparsify" blocks in Fig.~\ref{fig:fusion}.
As shown in our experiments in Sec.~\ref{sec:redundancy}, information is quite redundant within and across each modality, and we hypothesize simple sub-sequence pooling to be a cheap and effective method for capturing important information while removing redundancies.
Prior to pooling, we first apply a single bi-directional strided sparse attention layer \cite{child2019sparsetransformer} to enforce aggregation of dense local context and sparse global context to every token in the sequence to each modality.
We then apply non-overlapping per-channel pooling blocks of stride $s$ for each token set:
\begin{equation} \label{eq:pool}
    \tilde{\vect{z}}_{ij} = \mathrm{pool}\left(  \vect{z}^L_{i(js+1)}, \ldots, \vect{z}^L_{i(js+s)}\right).
\end{equation}
A natural choice for pooling is either per-channel max pool or average pool. We explored several options in ablation studies and found our method to be robust to the choice of pooling (see Table~\ref{tbl:pool}). However, for our main experiments we use average pooling.

We additionally form a multimodal classification token $\tilde{\vect{c}}$ by summing the unimodal classification tokens:
\begin{equation} \label{eq:cls-sum}
    \tilde{\vect{c}} = \sum_{i=1}^M \vect{c}_i^L
\end{equation}
The final, fused token set $\vect{F}$ is formed using this classification token and the union of the unimodal pooled token sets $\tilde{\vect{Z}}_1, \dots, \tilde{\vect{Z}}_M$:
\begin{equation}
    \vect{F} = [\tilde{\vect{c}}, \tilde{\vect{z}}_{11}, \dots, \tilde{\vect{z}}_{Mk}]
\end{equation}

\subsection{Dense Cross-modal Modeling and Prediction} \label{sec:cross-modal}
To model cross-modal relationships, we apply a dense, $T$-layer transformer on the token set $\vect{F}$.
Note the tokens of $\vect{F}$ are aggregated from all modalities.
We adopt the same architecture used in the unimodal modeling task, denoting the token set after $t$ transformer layers as $\vect{F}^t$, with the final output denoted $\vect{F}^T = [\tilde{\vect{c}}^T, \tilde{\vect{z}}_{11}^T, \dots, \tilde{\vect{z}}_{Mk}^T]$.
Finally, a small MLP followed by softmax is applied to $\tilde{\vect{c}}^T$ to produce a $C$-way class prediction $\vect{p}$.

\subsection{Multimodal Manifold Mixup} \label{sec:mixup}
We apply a novel variation of manifold mixup \cite{verma2018manifold} for improved generalization. 
In the originally proposed mixup \cite{zhang2017mixup}, given two random training inputs $\vect{x}_i$ and $\vect{x}_j$, their corresponding ground-truth labels $\vect{y}_i$, $\vect{y}_j$, and an interpolation weight $\lambda \in [0, 1]$, a classifier is trained using the following virtual training examples:
\begin{equation}
    \tilde{\vect{x}} = \lambda \vect{x}_i + (1-\lambda)\vect{x}_j
\end{equation}
\begin{equation} \label{eq:gt-mixup}
    \tilde{\vect{y}} = \lambda \vect{y}_i + (1-\lambda)\vect{y}_j
\end{equation}
Generally, the interpolation term $\lambda$ is sampled from a Beta distribution $\mathrm{Beta}(\alpha, \alpha)$, where $\alpha$ is a hyperparameter.
Manifold mixup extends this by also selecting a random layer $l$ in an $L$ layer network $f$ and interpolating the latent representations $\vect{v}_i^l, \vect{v}_j^l$ of that layer instead of the input example:
\begin{equation} \label{eq:manifold-mixup}
    \tilde{\vect{v}}^l = \lambda \vect{v}_i^l + (1 - \lambda) \vect{v}_j^l
\end{equation}
Layers $l+1, \dots, L$ of $f$ are then applied to $\tilde{\vect{v}}^l$ and the output is supervised using Eq.~\ref{eq:gt-mixup}.
Manifold mixup has been shown to be more effective for regularization than input mixup.

We extend manifold mixup to the multimodal case for use with our model.
Given our $(L+T)$-layer network, with the first $L$ layers involving separate, unimodal transformers and the last $T$ layers involving a single, multimodal transformer, we sample a single layer $l \in [1, L+T]$ for manifold mixup.
If $l > L$, we use standard manifold mixup using Eqs.~\ref{eq:gt-mixup} and~\ref{eq:manifold-mixup}.
If $l \leq L$, we sample a different interpolation term for each of the $M$ modalities, $\lambda_1, \dots, \lambda_M \sim \mathrm{Beta}(\alpha, \alpha)$.
Given latent representation $\vect{v}_{mi}^l, \vect{v}_{mj}^l$ of layer $l$ for modality $m$, the new latent representation is given as:
\begin{equation}
    \tilde{\vect{v}}_m^l = \lambda_m \vect{v}_{mi}^l + (1 - \lambda_m) \vect{v}_{mj}^l
\end{equation}
This is applied to every latent representation of layer $l$ for every modality $1, \dots, M$.
After running the remaining $L + T - l$ layers, the output of the network is supervised using:
\begin{equation}
    \tilde{\vect{y}} = \overline{\lambda_*} \vect{y}_i + (1 - \overline{\lambda_*})\vect{y}_j
\end{equation}
where $\overline{\lambda_*}$ is the average of the $M$ sampled $\lambda$ values.


\section{Experimental Setup}
We now describe the datasets used for training and evaluation (Sec.~\ref{sec:dset}), dataset pre-processing (Sec.~\ref{sec:preproc}), baseline network architectures used for comparison (Sec.~\ref{sec:nets}), and training hyper-parameters we used (Sec.~\ref{sec:hparams}).

\subsection{Datasets} \label{sec:dset}
We perform extensive experiments on two benchmark multimodal datasets: VGG-Sound \cite{chen2020vggsound} and CMU-MOSEI~\cite{zadeh2018multimodal} The datasets tackle popular and broadly applicable tasks in multimodal machine learning for audio-visual classification and multimodal sentiment classification. The modalities evaluated include video, audio, and text data. Additionally, these datasets have differences in modality characteristics such as cross-modality alignment and information content. 

\subsubsection{VGG-Sound} 
VGG-Sound \cite{chen2020vggsound} consists of over 200,000 YouTube videos and their associated audio streams, each annotated with one of over 310 class labels. The audio spans a large range of challenging acoustic environments and noise characteristics of real applications. All videos are captured ``in the wild." There are clear audio-visual correspondences, i.e., the sound source is visually evident. Each segment is 10 seconds long. To aid in evaluation, we select two subsets of data from VGG-Sound containing 10 classes and 100 classes each. We call these VGGS10 and VGGS100, respectively.
We select VGGS10 by choosing pairs of easily confused classes, such as ``baby babbling" and ``baby laughing". We then build VGGS100 using these ten classes and additionally include 90 randomly chosen classes. The total training and testing set sizes for VGGS10 are 6,051 and 459. For VGGS100, the training set size is 66,180 and the test set size is 4,549. A validation set is extracted by taking 20 percent of the training set.

\subsubsection{CMU-MOSEI} 
The CMU Multimodal Opinion Sentiment and Emotion Intensity (CMU-MOSEI) \cite{zadeh2018multimodal} dataset is one of the largest multimodal sentiment analysis and emotion recognition datasets to date. The dataset contains more than 23,500 sentence utterance videos from more than 1000 online YouTube speakers. The dataset is gender-balanced. All utterances are randomly chosen from various topics and monologue videos. The task is to predict a 7-class sentiment score of a particular multimodal video sample. Each sample contains audio, video, and text modalities. This dataset is frequently used to explore the unaligned nature of multimodal sequences between text and video. 


\subsection{Pre-processing} \label{sec:preproc}

Each modality is pre-processed with a feature extraction pipeline in order to generate the input token sequence. For the MOSEI dataset, we use the pre-processed data provided by the authors. The pre-processing pipeline that was used assumes that each video depicts a ``talking head": a single human talking, whose face is visible and whose voice is clearly audible. This assumption is valid for the MOSEI dataset, and the pre-processing pipeline therefore extracts visual features such as facial landmark positions and audio features such as estimated vocal parameters. We refer the reader to Zadeh \etal~\cite{zadeh2018multimodal} for the full details. To pre-process VGGSound, we employ a feature extraction pipeline that can be applied to videos more generally, without assuming human faces or voices are present.

For the VGGS10 and VGGS100 datasets, we extract \textit{visual features} using I3D \cite{carreira2017quo}, a spatio-temporal video feature extraction model that was pre-trained on the Kinetics human action recognition dataset~\cite{carreira2017quo}. This is a two-stream model, which processes optical flow and raw RGB independently as two separate modalities. We also extract TV-L$^1$ optical flow from the VGGSound videos. For \textit{Audio} pre-processing we follow Nagrani \etal~\cite{nagrani2021attention}: we resample all audio at 16Hz and convert to mono, then compute log mel spectrograms with 128 frequency bins, using a Hamming window with size 25ms and stride 10ms.



\subsection{Baseline Network Architectures} \label{sec:nets}


We compare against the following transformer-based fusion methods: 

\textbf{Self-Attention Fusion (Concat):} A baseline method of fusion is to concatenate the individual modality representations prior to input to any network and rely exclusively on dense self-attention. This is a form of early fusion.

\textbf{Late Fusion (LF):} This method works by applying transformer blocks on individual modalities only. The final prediction is obtained via a summation of logits derived from individual class tokens. This helps us compare the benefit of modeling cross-modal interactions. 

\textbf{Multimodal Transformer (MulT):}~\cite{tsai2019multimodal} MulT is a hybrid early-late attention-based fusion method using a unique cross-modal attention mechanism. The data is first fused via an attention mechanism by using one modality each for key, query, and value. Transformer blocks are then stacked on top. At the very end, the features are concatenated and a prediction is obtained after an FC layer.

\textbf{Bottleneck Fusion (MBT):}~\cite{nagrani2021attention} This is a form of fusion in which special tokens called bottleneck tokens are introduced. These tokens are shared among all modalities, and transformers alternate operating on each modality independently. The final \texttt{CLS} token is summed from each modality and used for prediction. We additionally evaluate MBT using manifold mixup (MBT+MM) as the original paper used input mixup, and our inputs are features.

\subsection{Implementation details} \label{sec:hparams}

Our model is implemented in PyTorch. For all experiments on the smaller datasets VGGS10 and MOSEI we use a learning rate of $10^{-4}$. For the larger dataset VGGS100 we use a learning rate of $10^{-3}$. Learning rate is decayed by factor of $10$ every 10 epochs based on minimum validation loss. We use a batch size of 24 for all experiments. For all datasets, we report results based on averaging performance training from 5 different seeds for generalization purposes and to minimize tuning effects. We use a standard 12-layer network and 5 attention heads for all evaluations. We project embeddings from each modality to 40 to minimize the effects of over-parameterization. For experiments involving latent mixup, we used a strength of $\alpha = 0.3$. We use an initial warm-up of 5 epochs in which no mixup is applied. For all other experiments we applied dropout $p = 0.2$ for regularization. For baselines, we follow descriptions in original papers and publicly available code for comparison. All experiments were conducted on consumer-grade graphics cards. We make our code and preprocessed data publicly available.

\subsection{Metrics}
We report results using commonly used metrics. \textbf{Top1} represents the accuracy of the most likely class. \textbf{mAP} represents the mean of per-class average precision scores. We also report the computational cost in Giga floating-point operations (GFlops) which is estimated similar to previous methods \cite{pan2021scalable} (we provide the equations used for estimating this in Appendix \ref{sec:flop_eq} included in the supplementary materials). Many experiments examine the effect of a reduction factor, which refers to reducing the number of tokens in the sequence dimension for transformer architectures. We report most results as a mean and standard deviation of experiments run with five different seeds.

\section{Results}

We first report our results against state of the art (Sec.~\ref{sec:sota}) showcasing our performance on multiple datasets from different domains. We then perform a series of ablation studies to explore the effects of sparsification (Sec.~\ref{sec:sparsification}), and the benefits of addressing within-modality redundancies during fusion (Sec.~\ref{sec:redundancy}). We also study the effect of pooling choice (Sec.~\ref{sec:pooling}) and the effect of our proposed multimodal manifold mixup (Sec.~\ref{sec:abl}).

\subsection{Comparison against state of the art}
\label{sec:sota}

\begin{table*}[t]
\small
\centering
\scalebox{0.89}{
\begin{tabular}{l|rr|rr|rr}
\toprule
 & \multicolumn{2}{c|}{\textbf{VGGS10}} & \multicolumn{2}{c|}{\textbf{VGGS100}} & \multicolumn{2}{c}{\textbf{MOSEI}} \\ 
 & \multicolumn{1}{c}{\textbf{Top1}} & \multicolumn{1}{c|}{\textbf{mAP}} & \multicolumn{1}{c}{\textbf{Top1}} & \multicolumn{1}{c|}{\textbf{mAP}} & \multicolumn{1}{c}{\textbf{Top1}} & \multicolumn{1}{c}{\textbf{mAP}} \\ \midrule
Concat &    $\underline{67.62} \pm 1.3$    &    $\textbf{71.46} \pm .63$ &    $51.72 \pm .26$   &  $51.64 \pm .13$   &        $48.47 \pm .23$   &  $\underline{32.40} \pm .83$      \\
LF     &     $67.10 \pm .79$    &  $70.46 \pm .79$            & $52.00 \pm .73$   &  $46.92 \pm .28$      &     $49.10 \pm .33$    &   $31.75 \pm .78$         \\
MulT   & $65.49 \pm .40$    &   $69.73 \pm 1.1$  &     $51.35 \pm .43$    &   $49.25 \pm .43$  &   $\underline{49.36} \pm .34$    &  $31.92 \pm .79$     \\
MBT    &  $66.84 \pm .61$    &   $70.98 \pm .78$  &    $51.67 \pm .66$ &  $51.29 \pm .37$     &       $49.12 \pm .27$    &  $32.15 \pm .47$        \\
MBT+MM &  $66.80 \pm 1.8$    &   $70.56 \pm .61$ &   $\textbf{55.97} \pm .42$    &   $\textbf{57.29} \pm .37$  & $48.77 \pm .37$    &   $32.03 \pm 1.2$  \\
Ours   &   $\textbf{67.71} \pm 1.3$   &  $\underline{71.06} \pm .81$  &  $\underline{55.61} \pm .61$    &   $\underline{57.18} \pm .39$  & $\textbf{49.67} \pm .23$ & $\textbf{33.66} \pm .85$  \\ \bottomrule         
\end{tabular}
}
\vspace{-0.2cm}
\caption{Accuracy comparison for each dataset and model. For all benchmarks we report the mean and standard deviation performance over 5 seeds to minimize tuning effects. Bold indicates best, underline second best. We are either best or close to best in all metrics.}
\label{tbl:main}
\end{table*}

\begin{table*}[t]
\centering
\small
\scalebox{0.89}{
\begin{tabular}{l|cccc|cccc}
\toprule
 & \multicolumn{4}{c|}{\textbf{VGGS10/VGGS100}}                                                                                             & \multicolumn{4}{c}{\textbf{MOSEI}}                                                                                                  \\
       & \multicolumn{1}{c}{\textbf{Mem (GB)}} & \multicolumn{1}{c}{\textbf{Eval (ms)}} & \multicolumn{1}{c}{\textbf{Train (ms)}} & \multicolumn{1}{c|}{\textbf{GFlops}} & 
       \multicolumn{1}{c}{\textbf{Mem (GB)}} & \multicolumn{1}{c}{\textbf{Eval (ms)}} & \multicolumn{1}{c}{\textbf{Train (ms)}} & \multicolumn{1}{c}{\textbf{GFlops}} \\
       \midrule
       
Concat & $1.52~(3.16\times)$                       & $3.59~(2.46\times)$ & $10.93~(2.56\times)$  & $1.68~(6.72\times)$                   & $1.04~(11.95\times)$ & $2.71~(2.39\times)$ & $8.02~(2.01\times)$  & $1.16~(11.60\times)$                     \\
LF     & $1.35~(2.80\times)$ & $3.54~(2.42\times)$ & $12.32~(2.88\times)$ & $1.51~(6.04\times)$  & $0.49~(5.66\times)$ & $2.00~(1.77\times)$ & $7.66~(1.92\times)$ & $0.59~(5.90\times)$     \\
MulT   & $1.18~(2.45\times)$ & $3.53~(2.41\times)$ & $16.73~(3.91\times)$  & $2.64~(10.56\times)$ & $0.62~(7.10\times)$ & $2.84~(2.50\times)$ & $11.49~(2.88\times)$  & $1.03~(10.30\times)$      \\
MBT    & $1.35~(2.82\times)$ & $3.59~(2.45\times)$ & $12.02~(2.81\times)$ & $1.52~(6.08\times)$ & $0.50~(5.72\times)$ & $2.14~(1.89\times)$ & $7.42~(1.86\times)$  & $0.59~(5.90\times)$                      \\
Ours   & \textbf{0.48}                         & \textbf{1.46}                          & \textbf{4.27}                           & \textbf{0.25}                       & \textbf{0.09}                         & \textbf{1.13}                          & \textbf{3.99}                           & \textbf{0.10}             \\
\bottomrule
\end{tabular}
}
\vspace{-0.2cm}
\caption{Computational cost comparison for each dataset and model. For all metrics we obtain results with a single RTX 3090. Metrics are normalized by the batch size. Our method has the lowest cost. GFlops is estimated based on number of transformer blocks and token operations and represents a theoretical cost for a single forward pass through the network. We present the equations used for calculations in the Appendix \ref{sec:flop_eq} of the supplementary material.}
\vspace{-.3cm}
\label{tbl:main_cost}
\end{table*}

We present our summary benchmark performance on real-world datasets VGGS10, VGGS100, and MOSEI in Tables~\ref{tbl:main} and~\ref{tbl:main_cost}. For each dataset, our model keeps a subset of tokens from each modality during pruning. For VGGSound data after pooling we have 12 tokens of RGB and flow information and 20 tokens spectrogram data. For MOSEI, we keep 10 tokens of visual and audio information and 25 tokens of text information. These numbers were chosen according to experiments described in Sec.~\ref{sec:redundancy}.

We maintain the performance of existing fusion methods and exceed them in some situations while significantly reducing the amount of computation required. For MOSEI we report more than a five-fold reduction in computational cost while achieving the best performance in terms of both Top1 accuracy and mAP. For VGGS10 and VGGS100, we observe approximately a six-fold reduction in computational cost. Our method also exceeds the performance of multiple fusion methods on the VGGS100 dataset.

\subsection{Effect of Sparsification}
\label{sec:sparsification}

\begin{table}[t]
\centering
\small
\scalebox{0.89}{
\begin{tabular}{@{}ll|rr|r@{}}
\toprule
& & \multicolumn{2}{c|}{\textbf{Token Reduction Factor}}    & \\
                             &      & \textbf{None} & \textbf{64$\times$} & \textbf{Diff.} \\\midrule
\multirow{2}{*}{\textbf{Concat}}      & \textit{Top1} &     $51.72 \pm .26$     &       $46.29 \pm .73$              &      {\color{red} $-5.45$}      \\
                             & \textit{mAP}  &      $51.64 \pm .13$        &        $47.49 \pm .63$             &      {\color{red} $-4.49$}      \\ \midrule
\multirow{2}{*}{\textbf{LF}} & \textit{Top1} &      $52.00 \pm .73$        &  $49.76 \pm .71$               &     {\color{red} $-2.24$}       \\
                             & \textit{mAP}  &     $46.92 \pm .28$         &        $45.96 \pm .62$             &     {\color{red} $-0.96$}      \\\midrule
\multirow{2}{*}{\textbf{Ours}}        & \textit{Top1} &      $\textbf{55.57} \pm .23$        &      $\textbf{55.98} \pm .28$     &       {\color{teal}$+0.41$}     \\
                             & \textit{mAP}  &      $\textbf{56.54} \pm .49$        &         $\textbf{56.91} \pm .60$    &   {\color{teal}$+0.37$}   \\ \bottomrule
\end{tabular}
}
\vspace{-0.2cm}
\caption{Comparison of our method for sparsification versus application of only pooling in baseline methods on VGGS100. \texttt{Diff} column shows difference between no reduction of tokens and taking 1/64ths of the tokens, where the minimum is one token per modality. Our method is more robust than naive methods of pooling. Pooling has a large effect when training with fused features (Concat) which we solve using our method. Difference for the same reduction factors between Top1 and mAP shows that late fusion (LF) tends to fit some samples better than others and suggests the advantages of an early-fusion method.}
\label{tab:sparsification}
\end{table}

\begin{figure*}[t]
     \centering
     \begin{subfigure}[t]{.32\linewidth}
         \centering
         \includegraphics[width=.95\linewidth]{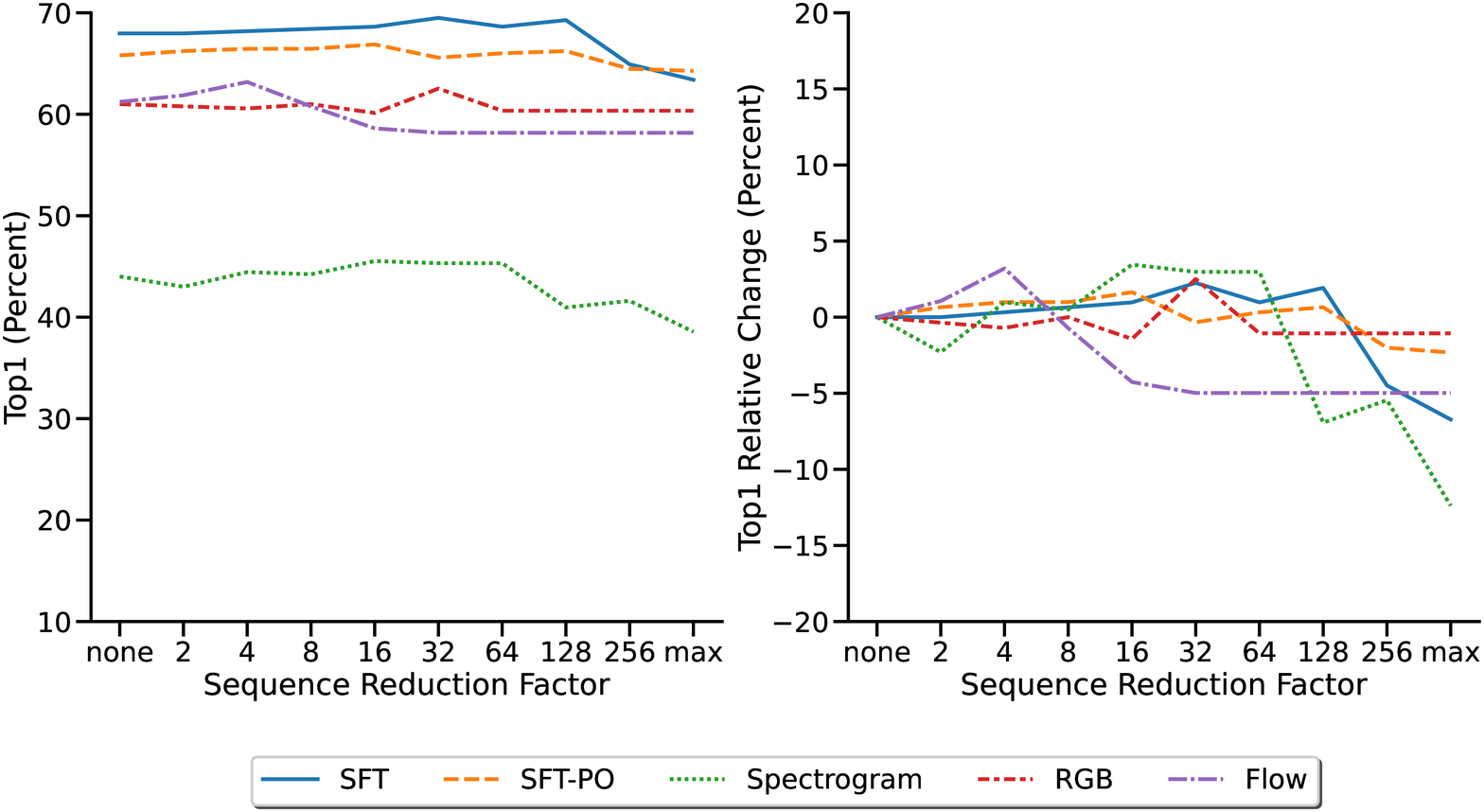}
         \caption{Top 1 absolute score and relative change from no pooling for VGGS10. Multimodal performance degredation occurs after a 64-fold reduction in sequence length. Compared to flow at 4, and spectrogram at 64. We outperform all all methods at all reduction levels. SFT exceeds the pooling only variant (SFT-PO).}
         \label{fig:reduction_vgg10}
     \end{subfigure}
     \hfill
     \begin{subfigure}[t]{.32\linewidth}
         \centering
         \includegraphics[width=.95\linewidth]{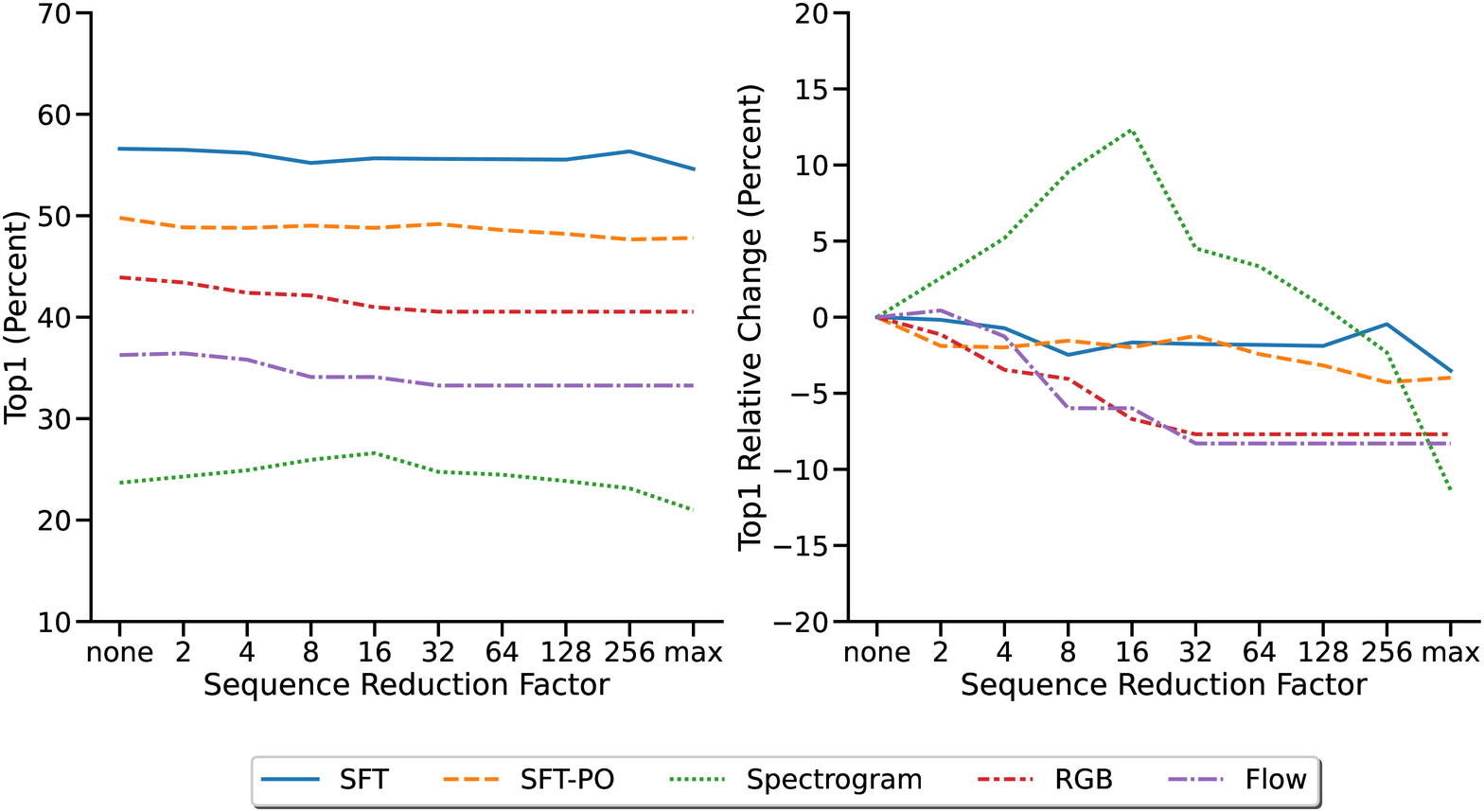}
         \caption{Top 1 metrics for VGGS100. SFT degradation occurs at 256-fold reduction compared to 64 for SFT-PO and 2 for Flow and RGB. Audio representations might benefit from better feature extraction, however there is dramatic loss of performance with very few tokens, while we remain tolerant.}
         \label{fig:reduction_vgg100}
     \end{subfigure}
     \hfill
     \begin{subfigure}[t]{.32\linewidth}
         \centering
         \includegraphics[width=.95\linewidth]{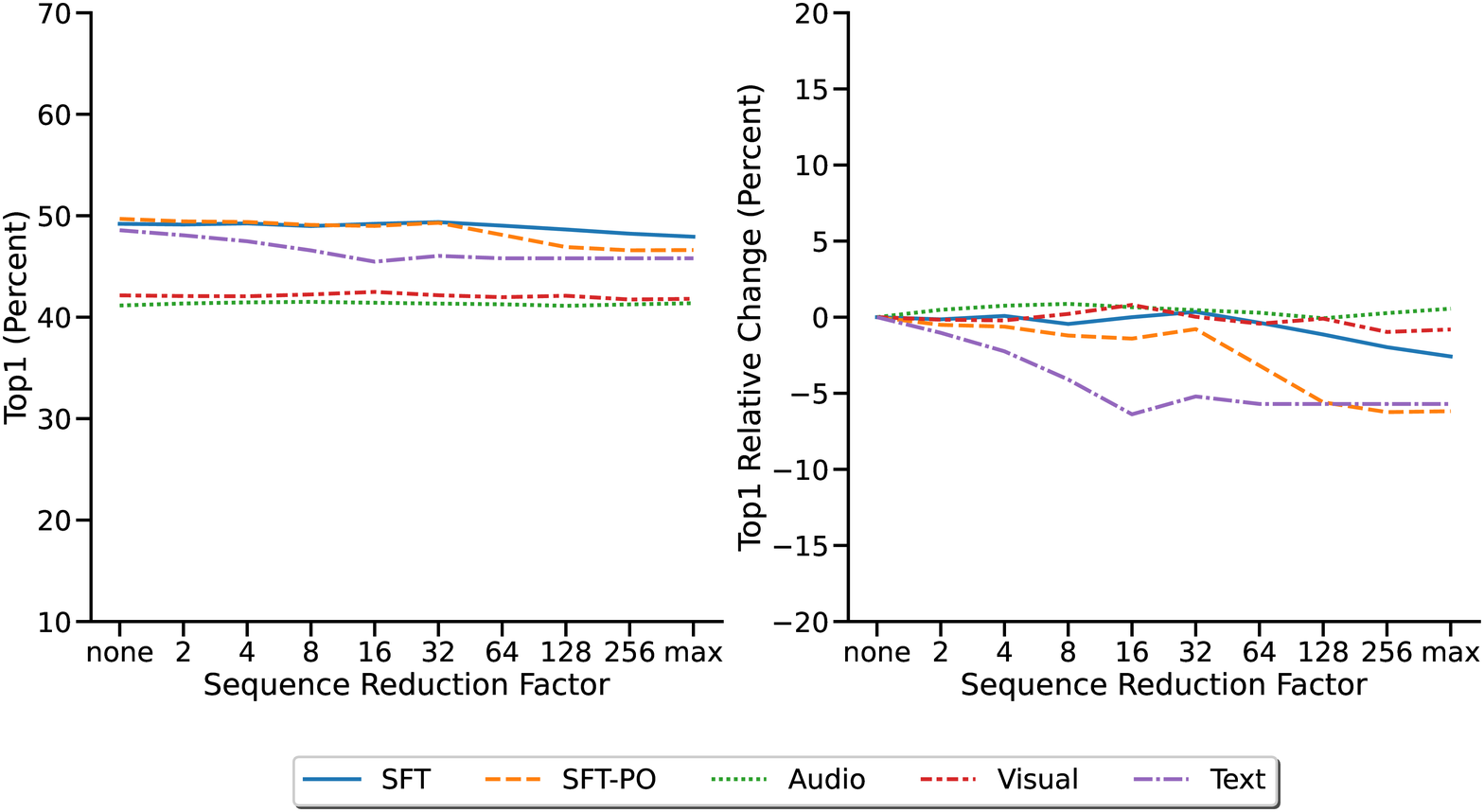}
         \caption{Top 1 metrics for MOSEI. SFT degrades minimally until max while SFT-PO degrades at 32. Text modality degrades immediately. Information appears highly redundant in Audio-Visual modalities.}
         \label{fig:reduction_mosei}
     \end{subfigure}
    \caption{Comparison of reduction factor effect on performance difference against no reduction for unimodal and multimodal models. Reduction in total length of fused features reported in the x-axis. In cases where the reduction factor is greater than sequence length of a particular modality, a single token along the sequence dimension is passed through. Sequence lengths for VGGS10 and VGGS100 are 38 for RGB and Flow, 1200 for Spectrogram. For MOSEI, Audio and Visual is 500 while text is 50. Top1 absolute score and relative change from using no pruning is reported. For all experiments we used a batch size of 24. Multimodal models will tolerate more pruning over unimodal models by making up for the lost information through fusion. Notably, SFT exceeds performance of SFT without sparse attention or mixup (SFT-PO) in all cases and tolerates more reduction. Pooling offers some benefits for feature extraction in some cases for longer sequences.}
    \vspace*{-3mm}
    \label{fig:reduction_factor}
\end{figure*}

In this section, we explore the effect of how naively applying pooling can affect multimodal models. In particular, we are interested in how pooling affects fused versus modality-independent features. We answer this question by comparing the performance of late fusion, concatenation fusion, and our fusion method. For concatenation fusion, we concatenate all the input tokens prior to input into the model. From here, we apply a single transformer block as if the number of modalities is $M=1$. We then apply max pool with a kernel and stride of 64. Afterwards, we apply eleven more transformer layers to obtain the result. For late fusion and our method, we also apply pooling on the representations after the first layer. However, the pooling is conducted on unimodal representations. In late fusion, transformer layers are applied independently for each modality and the final result is obtained via a summation of logits obtained from the \texttt{CLS} token. In experiments described in Sec.~\ref{sec:redundancy}, we observe a drop in performance for our method with strides larger than 32 for some datasets and 128 for others, thus we assume a stride of 64 will provide meaningful comparisons between fusion methods.

The results shown in Table \ref{tab:sparsification} demonstrates that our method for sparsification is more robust than naive methods. We see that in both naive methods of pooling, the reduction in the sequence dimension causes a significant drop in performance. Our method does not see any reduction, instead experiencing a small boost in performance. Furthermore, we see that concatenation fusion tends to have a higher mAP metric, whereas late fusion has a higher Top1. Overall, our method is robust, and pooling has no detrimental effect even when removing over $98\%$ of tokens.

\subsection{Within-Modality Information Redundancy}
\label{sec:redundancy}

We provide experiments to analyze why it is advantageous to address the within-modality redundancy problem during fusion. In particular, we wish to show that pooling when accounting for multimodal information is more robust than pooling without this information. We set up the experiment so that a max-pooling layer is applied after the first layer of transformers to simulate modality-independent feature sparsification for each method. We then compare pruning by an equal factor for each modality to observe the effect on overall performance, referred to as ``sequence reduction factor." We set the minimum allowed sequence length to one to avoid removing all tokens. We compare against unimodal transformers for each modality. We also evaluate two versions of our method: SFT which is our full pipeline, and SFT-PO which removes the strided sparse attention layer and multimodal manifold mixup and includes only the strided pooling.

In the first column of Fig.~\ref{fig:reduction_factor}, we present Top1 accuracy as a function of sequence reduction factor.
In the second column, we present the relative change in Top1 accuracy when compared with no sequence reduction. Lower indicates a performance degradation from sequence reduction.
Multimodal models exceed unimodal performance in all reduction factors. We generally see a performance decrease for each unimodal model as the reduction factor increases. However, some modalities do not decrease due to two likely reasons: 1) from redundancies in information and 2) that all useful information was extracted after just a single layer of transformers. We also see that some modalities experience an increase in performance as we reduce the number of tokens, signifying better feature extraction for those. However, in general, the performance of unimodal models with less redundant information all decrease, while our model (SFT) is more robust. In particular, SFT is better than using just pooling (SFT-PO) as is evident from it maintaining higher performance with greater reduction factors.

\begin{table*}
\centering
\small
\scalebox{0.89}{
\begin{tabular}{@{}l|ll|ll|ll} 
\toprule
\multicolumn{1}{r|}{\multirow{2}{*}{\textbf{Pooling Method}}} & \multicolumn{2}{c|}{\textbf{VGGS10}}                                    & \multicolumn{2}{c|}{\textbf{VGGS100}}                                   & \multicolumn{2}{c}{\textbf{MOSEI}}  \\                                
\multicolumn{1}{r}{}                                         & \multicolumn{1}{|c}{\textit{Top1}} & \multicolumn{1}{c}{\textit{mAP}} & \multicolumn{1}{|c}{\textit{Top1}} & \multicolumn{1}{c}{\textit{mAP}} & \multicolumn{1}{|c}{\textit{Top1}} & \multicolumn{1}{c}{\textit{mAP}} \\ 

\midrule
Max & 67.0 $\pm$ 1.1 & 70.7 $\pm$ 0.7 & 55.7 $\pm$ 0.4 & 57.3 $\pm$ 0.4 & 49.4 $\pm$ 0.2 & 33.2 $\pm$ 0.3 \\
Average& 67.7 $\pm$ 1.2 & 71.1 $\pm$ 0.7 & 55.6 $\pm$ 0.5 & 57.2 $\pm$ 0.3 & 49.7 $\pm$ 0.2 & 33.7 $\pm$ 0.9 \\
Attn Average& 67.5 $\pm$ 1.0 & 71.1 $\pm$ 0.7 & 55.4 $\pm$ 0.3 & 56.9 $\pm$ 0.4 & 49.4 $\pm$ 0.2 & 33.7 $\pm$ 0.8 \\
\bottomrule
\end{tabular}
}
\vspace{-0.2cm}
\caption{Comparison of pooling method on VGGS10, VGGS100, and MOSEI datasets. Based on Top 1 accuracy and mean average precision metrics, we find our method robust to pooling type.}
\vspace{-.3cm}
\label{tbl:pool}
\end{table*}

We see that up to a factor of 50 for evaluations conducted on MOSEI, there is very minimal drop in performance in the multimodal model. However, the performance of the text-only transformer drops observably larger than our multimodal model. The performance of the RGB and Audio transformers remains the same throughout the experiment. This signifies two things: that the information for label present in the text classifier is less redundant than in RGB and Audio features for this dataset, and that application of sparse fusion can compensate for the loss of information necessary for classification by exploiting the other modalities. The effect of unimodal models experiencing a decrease in performance is also evident for the optical flow modality on the VGGS10 dataset at 8$\times$ reduction, and at 64$\times$ reduction for spectrogram data. On VGGS100, we see the same, where both the RGB and flow modalities experience decreases in performance with a pruning factor of just 2$\times$ while our model's performance remains relatively flat. Furthermore, our multimodal model with one token per-modality after pruning still achieves better performance than a unimodal model which uses all tokens. 

These observations signify that certain modalities contain information that is more redundant than others and that even if we filter out more than what a model with redundant information is able to predict, the multimodal model is able to make up for that. The same is not true for unimodal models, which cannot filter out unnecessary information as well, and is not robust to this reduction. Even under extreme circumstances where information is reduced to the length of a single token, performance of the multimodal degrades but still remains the overall top performer.

\subsection{Effect of Pooling}
\label{sec:pooling}

In this section, we explore the effects of using various pooling choices in the network.
See Table~\ref{tbl:pool} for results on the VGGS10, VGGS100, and MOSEI datasets.
We use max pooling, average pooling, and attention-weighted average pooling, denoted ``Max," ``Average," and ``Attn Average" respectively.
For attention-weighted averaging, we weight using a simple, attention-based per-token significance metric proposed by Goyal \etal~\cite{goyal2020powerbert}.
Given the attention weights $\vect{W}^h \in \mathbb{R}^{N \times N}$ calculated from layer $L$ head $h \in \{1, \dots, H\}$ of the pre-fusion network, the significance (sig) for token $i$ is:
\begin{equation} \label{eq:sig}
    \mathrm{sig}(i) = \sum_{h=1}^{H} \sum_{n=1}^{N}{W}^{h}_{in}
\end{equation}
Interestingly, all metrics are within 1 percentage point of each other across the three pooling types. This indicates our model is quite robust to the choice of pooling type. Average pooling appears better, but this is well within the std.~dev.





\subsection{Effect of Multimodal Manifold Mixup}
\label{sec:abl}

\begin{table}
\small
\centering
\begin{tabular}{@{}c|rr@{}}
\toprule
  mixup? & \textbf{Top1} & \textbf{mAP} \\ \midrule
\checkmark & $55.61 \pm .61$ & $57.18 \pm .39$\\
$\times$ & $51.30 \pm .80$ & $51.80 \pm .44$ \\
\bottomrule
\end{tabular}
\vspace{-0.2cm}
\caption{Comparison of model performance on VGGS100 when trained with and without our multimodal manifold mixup.} 
\label{tab:mixup}
\end{table}

See Table~\ref{tab:mixup} for results from SFT trained on VGGS100 with and without the use of our multimodal manifold mixup during training. Without mixup, we observe over a $4\%$ reductin in Top1 and over a $5\%$ reduction in mAP. This drop in performance is quite significant, indicating the effectiveness of training with our proposed multimodal manifold mixup.

\section{Limitations}
We provide an effective method for quickly ingesting and classifying large quantities of multimodal sequential data with high levels of accuracy. However, we do not provide evaluations on how this fusion method might behave as part of a generative network and we leave this for future work. Secondly, our methods operate on extracted features such as I3D and spectrogram data. While we follow popular and common settings for feature extraction, improved unimodal modeling might be able to condense the representations and reduce within-modality redundancy. This would lead to slightly reduced complexity benefits. However, the large differences between our results and unimodal approaches as well as maintaining performance under extreme sparsification support our conclusions.

\section{Conclusion}
We present an effective technique that offers more than a five-fold reduction in computational cost while maintaining the performance of state-of-the-art fusion techniques. Different fusion methods exhibit improved performance under varying conditions when all input conditions are equal. However, when optimizing for speed, there are drastic improvements that can be made to feature selection during cross-modal modeling that can improve performance.

\textbf{Broader Impacts:}
We propose sparse fusion for multimodal transformers as a method to reduce computational costs. This translates to energy savings and is beneficial for numerous applications including on mobile devices. Namely, it has the potential to train and fine-tune a network for use to a specific user without needing to offload the training to a server. This preserves the privacy of the user while providing benefits of performance and energy savings. Furthermore, we hope to spur democratization of learning on large datasets by enabling rapid development and evaluation on consumer-level hardware. However, we hope that by enabling this technology on mobile devices it is not applied to tasks such as unlawful surveillance.

{\small
\bibliographystyle{ieee_fullname}
\bibliography{main.bib}
}

\clearpage
\appendix

\section{Appendix}

\subsection{Label Distributions}
\label{sec:lbl_dist}
\begin{figure}[ht]
\centering
\includegraphics[width=.95\linewidth]{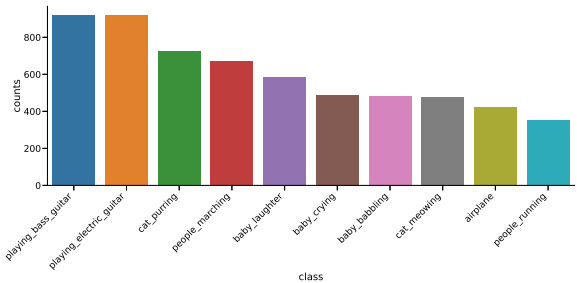}
\caption{Label distribution of VGGS10 dataset}
\label{fig:vggs10_dist}
\end{figure}

\begin{figure}[ht]
\centering
\includegraphics[width=.95\linewidth]{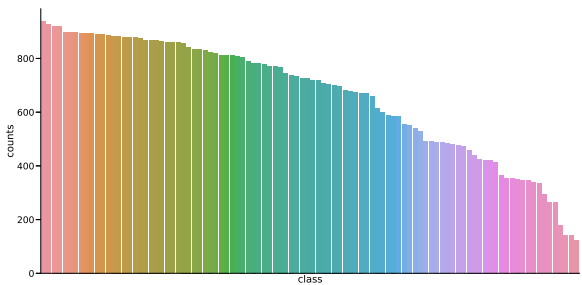}
\caption{Label distribution of VGGS100 dataset}
\label{fig:vggs100_dist}
\end{figure}

We summarize the classes we used in VGGS10 and provide the label distributions in VGGS10 and VGGS100. VGGS10 is a manually curated dataset built by selecting pairs of difficult to separate classes from the full VGGSound dataset as well as for differences between video and audio modalities. We chose the following ten classes: airplane, baby babbling, baby crying, baby laughter, cat meowing, cat purring, people marching, people running, playing bass guitar, playing electric guitar. The final training set distribution for VS10 in Fig. \ref{fig:vggs10_dist}, and the final VGGS100 dataset distributions are show in Fig. \ref{fig:vggs100_dist}.

\subsection{Flop computation}
\label{sec:flop_eq}

We present all flop estimates in the paper using the following equations. We primarily follow the flop estimation from \cite{pan2021scalable} with some minor changes due to layer differences. Each transformer layer consists of a multi-head attention and multilayer perceptro block. A multi-head attention (MHA) block has cost of:
\begin{align}
\phi_{MHA} =& ~\phi_{qkv} + \phi_A + \phi_O + \phi_{proj} \nonumber \\
=&~3nd^2 + n^2d + n^2d + nd^2 \nonumber \\
=&~4nd^2 + 2n^2d
\end{align}
where $n,d$ represent the length and embedding dimension, $\phi_{qkv}$ is the cost of projecting to the query, key, and values. $\phi_{A}$ is the cost of the attention map, $\phi_{O}$ is the cost of the self attention, and $\phi_{proj}$ is the cost of projection for self-attention outputs.

A MLP block includes two linear layers as well as a normalization layer for a cost of:
\begin{align}
    \phi_{MLP} =&~\phi_{proj1} + \phi_{norm} + \phi_{a} + \phi_{proj2} \nonumber \\
     =&~nd^2 + 3nd + nd + nd^2 \nonumber \\
     =&~2nd^2+4nd
\end{align}
where $\phi_{proj1}$ and $\phi_{proj2}$ are cost of projecting into and out of latent space for transformer block, $\phi_{norm}$ represents cost of applying layer normalization, and $\phi_{a}$ represents the cost of an activation function.

\end{document}